\documentclass[10pt,twocolumn,letterpaper]{article}

\usepackage[pagenumbers]{cvpr} % To force page numbers, e.g. for an arXiv version

\definecolor{cvprblue}{rgb}{0.21,0.49,0.74}
\usepackage[pagebackref,breaklinks,colorlinks,allcolors=cvprblue]{hyperref}

\def\paperID{} % *** Enter the Paper ID here
\def\confName{CVPR}
\def\confYear{2026}

\newcommand{\ours}{PAD-Hand}
\title{\ours: Physics-Aware Diffusion for Hand Motion Recovery}

\author{
Elkhan Ismayilzada\textsuperscript{1} \qquad
Yufei Zhang\textsuperscript{2} \qquad
Zijun Cui\textsuperscript{1} \\
\\
\textsuperscript{1}Michigan State University \qquad \textsuperscript{2}Independent Researcher\\
{\tt\small \{ismayil1, cuizijun\}@msu.edu, yufeizhang96@outlook.com}
}
\usepackage{multirow}
\usepackage{booktabs} 
\usepackage{algorithm,algpseudocode}

\newcommand{\yt}{y_{1:T}}
\newcommand{\xt}{x_{1:T}}
\newcommand{\hatxt}{\hat{x}_{1:T}}
\newcommand{\noi}{\kappa}
\newcommand{\alpsche}{\alpha}
\newcommand{\etasche}{\eta}
\newcommand{\res}{d_{1:T}}

\begin{document}

\twocolumn[{%
	\renewcommand\twocolumn[1][]{#1}%
	\maketitle
	\begin{center}
		\newcommand{\teaserwidth}{\textwidth}
		\centerline{
			\includegraphics[width=\teaserwidth,clip]{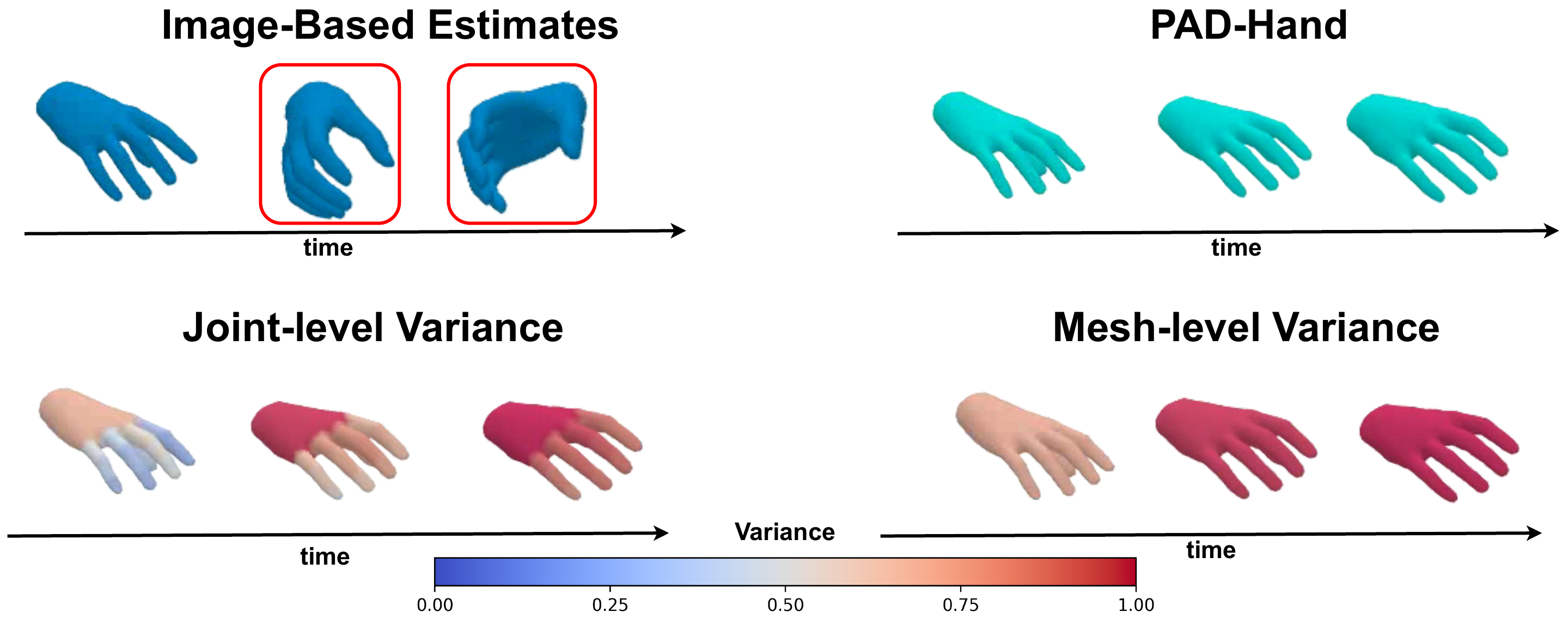}
		}
		\captionof{figure}{\textbf{Refined motion estimates by PAD-Hand with dynamic variance.} \textbf{Top}: Image-based estimates (left) are refined by our model (PAD-Hand) (right) to enforce temporal and physics consistency.
\textbf{Bottom}: Joint-level (left) and mesh-level (right) variance maps concentrate on frames/regions where the image-based motion estimate is unreliable (highlighted in red), aligning high variance with poor motion estimates. The color bar shows normalized variance (low to high).}
		\label{fig:teaser}
	\end{center}%
}]

\maketitle
\begin{abstract}
Significant advancements made in reconstructing hands from images have delivered accurate single-frame estimates, yet they often lack physics consistency and provide no notion of how confidently the motion satisfies physics. In this paper, we propose a novel physics-aware conditional diffusion framework that refines noisy pose sequences into physically plausible hand motion while estimating the physics variance in motion estimates. Building on a MeshCNN–Transformer backbone, we formulate Euler–Lagrange dynamics for articulated hands. Unlike prior works that enforce zero residuals, we treat the resulting dynamic residuals as virtual observables to more effectively integrate physics. Through a last-layer Laplace approximation, our method produces per-joint, per-time variances that measure physics consistency and offers interpretable variance maps indicating where physical consistency weakens. Experiments on two well-known hand datasets show consistent gains over strong image-based initializations and competitive video-based methods. Qualitative results confirm that our variance estimations are aligned with the physical plausibility of the motion in image-based estimates. 
\end{abstract}  
\section{Introduction}
\label{sec:intro}
Recovering hand motions from monocular cameras aims to reconstruct a hand’s 3D configuration over time from single-view images. This is essential for applications such as augmented and virtual reality (AR/VR) \cite{chen2023hand, pei2022hand, dong2024physical, zhao2021m3d}, human–computer interaction \cite{ren2020review, huo20233d}, and beyond.

Following the advancement of deep learning, substantial progress has been made in recovering hand meshes from monocular images \cite{tse2022a, yang2022a, cho2023transformer, aboukhadra2023thor, ismayilzada2025qort, kim2021a, meng2022a, moon2023a, ren2023a, wang2023a, yu2023a, zhang2021a, zuo2023a, zhang2024weakly} or videos \cite{fu2023deformer, ziani2022tempclr, baradel2022posebert, zhou2022toch}. Particularly, scaling up to build foundation models has brought continued performance gains~\cite{pavlakos2024reconstructing, potamias2024wilor, dong2024hamba}, although these models remain solely image-based due to the cost of acquiring video data. They improve per-frame accuracy, yet temporal inconsistency persists. Moreover, all these approaches primarily capture hand kinematic patterns and remain agnostic to dynamics, which suffers from the physical implausibility and limits their applicability in scenarios such as embodied AI.

To address these challenges, incorporating physics into 3D hand reconstruction represents a promising direction. Some works integrate physics through external simulators in related domains such as full-body motion and hand–object interaction \cite{li2019estimating, hu2022physical, braun2024physically, shimada2024macs, hu2024hand, xie2024ms}. They obtain initial motion estimates from kinematics-based models and then learn control policies or impose explicit physical constraints to refine the results in a post-hoc manner. Other works instead effectively leverage deep learning and directly build physics-informed models for motion refinement~\cite{zhang2024physpt, luo2024physics}. The most recent work by Zhang \textit{et al.}~\cite{zhang2025diffusion} achieves further improvements by capturing the uncertainty in initial motion estimates using diffusion models. Nonetheless, a key limitation of existing physics-based methods lies in their deterministic incorporation of physics. They enforce physics losses by driving equation residuals to zero rather than modeling a distribution over these residuals. Such approaches assume the observed motion can satisfy physics exactly, with no noise, bias, or model mismatch. However, 3D hand estimates are noisy and physics models are approximate, so forcing zero residuals ignores inherent uncertainty. 
This perfect-data assumption can lead to a challenging optimization landscape~\cite{krishnapriyan2021characterizing} or suboptimal solutions~\cite{ramirez2025bayesian}. Deterministic formulations also fail to reveal unreliable cases, limiting interpretability. This motivates a probabilistic formulation of physics incorporation, enabling the model to reason over motion data manifold and shape a distributional solution space.

In this work, we introduce a novel physics-aware diffusion framework that effectively incorporates hand dynamics for 3D hand motion recovery from video. We formulate Euler–Lagrange (EL) dynamics for articulated hands and incorporate it into diffusion models. We treat the resulting dynamic residuals as virtual observables drawn from a distribution. Their likelihood is coupled with the standard visual data terms to guide the reverse diffusion process toward physically admissible trajectories. To obtain interpretable physical consistency, we apply a last-layer Laplace approximation to measure variances over associated dynamics residuals. The estimated variances serve as an indicator of uncertainty in the recovery process, allowing us to identify potentially unreliable segments, as shown in Figure~\ref{fig:teaser}. In summary, our contributions are as follows:
\begin{itemize}
  \item \textbf{A probabilistic framework for physics integration.} We cast Euler--Lagrange residuals as \emph{virtual observables}~\cite{rixner2021probabilistic, bastek2024physics} and integrate their likelihood into the diffusion objective, enabling a probabilistic integration of physics.
  \item \textbf{Interpretable physics consistency.} We estimate variances of dynamic residuals through a last-layer Laplace approximation, which provides a principled indicator of recovery uncertainty and helps interpret where physical consistency may weaken.
  \item \textbf{Comprehensive evaluation.} On two widely adopted hand datasets, we improve reconstruction accuracy  and physical plausibility. Qualitative results confirm that our variance estimations are aligned with the physical plausibility of the motion in image-based estimates.
\end{itemize}

\section{Related Works}
\begin{figure*}[t!]
    \centering
    \includegraphics[width=\textwidth]{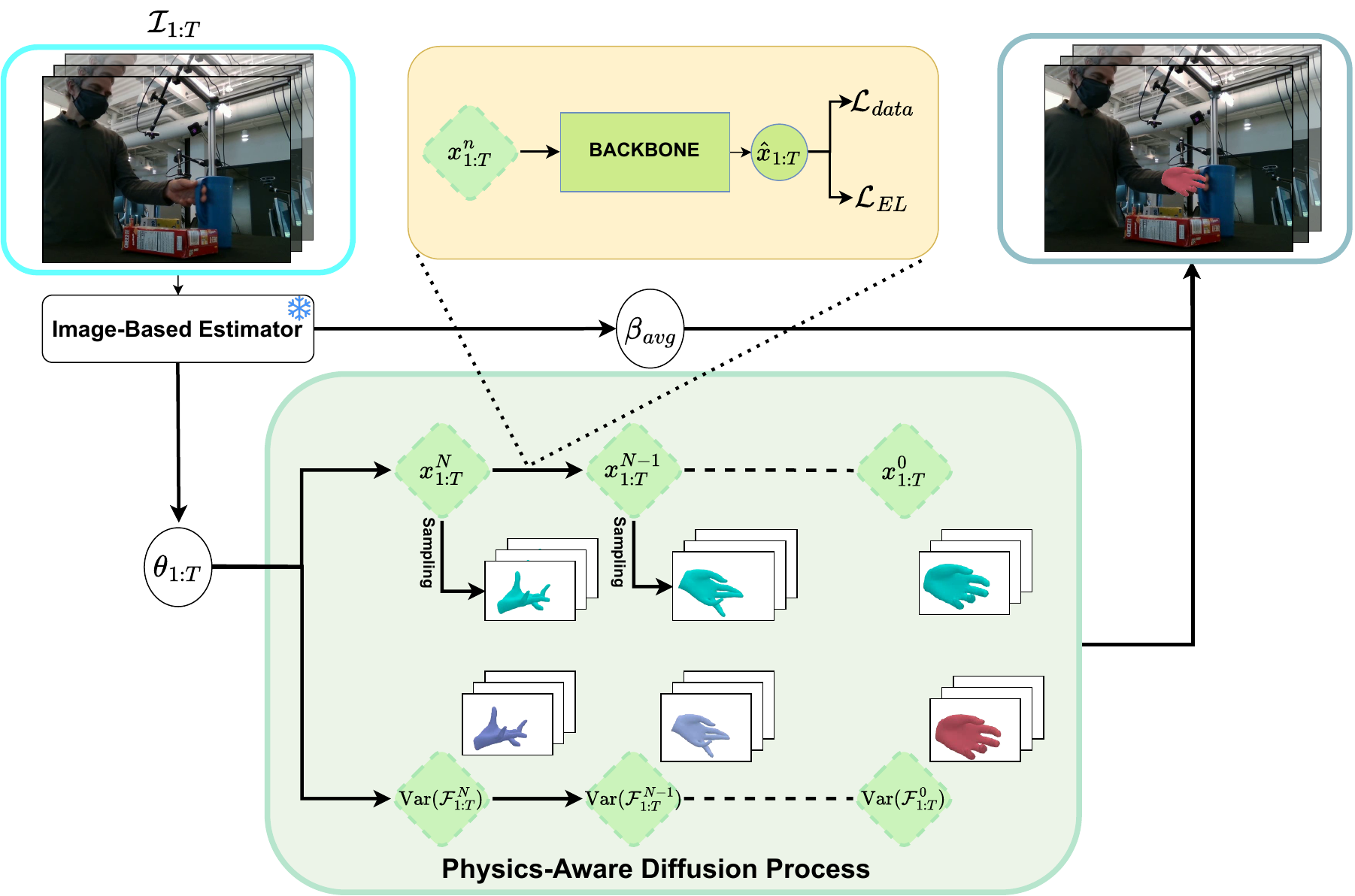}
    \caption{\textbf{Overview of PAD-Hand.} A sequence of images $\mathcal{I}_{1:T}$ is passed
through an image-based pose estimator to obtain per-frame pose
$\theta_{1:T}$ and the average shape $\beta_{avg}$ estimates. The pose estimates are then refined via a diffusion process to obtain temporally coherent motion. Simultaneously, we propagate the variance at each diffusion step starting from delta Dirac distribution at diffusion step $N$  to obtain per-frame dynamic variance estimates. At each diffusion step, backbone predicts clean motion $\hatxt$ which is supervised with data-driven loss $\mathcal{L}_{data}$ and physics-driven loss $\mathcal{L}_{EL}$ during training. }
    \label{fig:diffusion}
\end{figure*}
\noindent\textbf{Monocular 3D Hand Reconstruction.}
Current methods for reconstructing 3D hands from images~\cite{baek2019a,boukhayma2019a,zhang2019a,chen2022mobrecon,zimmermann2017a,jiang2023a2j,zhang2024handformer2t,pavlakos2024reconstructing,zhou2024simple, zhang2024weakly} achieve strong single-frame accuracy and have proven effective in specialized scenarios such as hand–object \cite{hasson2019a,hasson2020a,hampali2020a,tse2022a,yang2022a,cho2023transformer,aboukhadra2023thor,ismayilzada2025qort} and hand–hand interactions \cite{kim2021a,meng2022a,moon2023a,ren2023a,wang2023a,yu2023a,zhang2021a,zuo2023a}. Recent models~\cite{potamias2024wilor,pavlakos2024reconstructing,dong2024hamba} trained on larger models and datasets further improve generalization. To further promote temporal consistency, other methods take video input and employ temporal contrastive learning \cite{ziani2022tempclr} or temporal encoders \cite{fu2023deformer} to capture motion dependency. Due to scarcity of high-quality annotated hand video datasets, these methods may still fail to outperform image-based approaches in per-frame reconstruction accuracy. Instead, some works \cite{baradel2022posebert, zhang2024physpt, luo2024physics, zhang2025diffusion, zhou2022toch} tackle this limitation by recovering hand motion from image-based estimates. These methods are trained solely on 3D motion capture data without videos, providing strong motion priors that encourage temporal consistency. However, consistency alone is not a sufficient factor for a physically plausible hand motion. Such approaches provide data-driven motion priors ignoring the adherence to the governing laws of motion dynamics.

\noindent\textbf{Physics-Informed Hand Motion Modeling.} Existing efforts to incorporate physics largely target hand–object interactions. Some works \cite{christen2022d, yang2022learning} simulate interactions in virtual environments, using physics-informed deep reinforcement learning to obtain realistic grasping and contact patterns. A common limitation of these approaches is their dependence on physics simulators, which  complicates end-to-end learning. Another line of work \cite{hu2022physical, braun2024physically, shimada2024macs, hu2024hand} leverages object information to enforce physical constraints during hand reconstruction, but these methods assume the availability of accurate object geometry and trajectories. They also remain deterministic, offering no mechanism to capture or report predictive uncertainty. More recently, Zhang \textit{et al.}~\cite{zhang2025diffusion} employ a conditional diffusion model regularized by intuitive physics terms to encourage natural hand motion. Although conditioning on motion states derived from hand–object proximity can improve realism near contact, this design is inherently object-dependent and provide insufficient information about hand dynamics, such as motion forces. Moreover, existing methods lack a proper mechanism to incorporate physics when the observed motion does not satisfy physics exactly.

\noindent\textbf{Probabilistic Diffusion Model.} Several works in the literature propose ways to incorporate the physics into diffusion models \cite{rixner2021probabilistic, jacobsen2025cocogen, bastek2024physics, shu2023physics}. Rixner \textit{et al.} \cite{rixner2021probabilistic} propose to integrate physics residuals as virtual observables and compute ELBO approximation over the joint distribution of the data and the residuals.  Bastek \textit{et al.} \cite{bastek2024physics} simplify this process by treating physical likelihood as an auxiliary training objective. However, none of these works propose a way to estimate the variance of their estimations. While several methods quantify uncertainty in image generation~\cite{kou2023bayesdiff,de2025diffusion}, such as Kou et al.~\cite{kou2023bayesdiff} who apply a last-layer Laplace approximation to obtain pixel-wise uncertainty, we are the first to incorporate physics probabilistically and estimate variance in 3D hand motion recovery.

\section{Method}

\label{sec:method}
\textbf{Overview}. An overall of our proposed framework is shown in Figure \ref{fig:diffusion}. Our pipeline takes a sequence of $T$ frames $\mathcal{I}_{1:T}$ as input and first runs an off-the-shelf hand-pose estimator to obtain per-frame MANO~\cite{romero2017a} shape $\beta_{1:T}$ and pose $\theta_{1:T}$ estimates. Since hand shape remains constant throughout the sequence, we consider the mean shape $\beta
_{avg}=\frac{1}{T}\sum_{i=1}^T\beta_i$. Then, we introduce a conditional diffusion model that iteratively refines the motion from a noisy state $x_{1:T}^N$ to a clean state $x_{1:T}^0$ over $N$ diffusion steps, conditioned on the mean shape and per-frame pose estimates $y_{1:T}=\{\beta_{avg},\theta_{1:T}\}$. Simultaneously, our model produces per-step dynamic variances $\mathrm{Var}(\mathcal{F}_{1:T}^n)$ for $n\in[1,N]$. The final output is the refined pose trajectory $x_{1:T}^0$ together with the variance estimates of associated dynamics residuals $\text{Var}(\mathcal{F}^0_{1:T})$.
In Section \ref{sec::el}, we formulate the Euler-Lagrange equation for hand motion, and in Section \ref{sec::diffusion}, we introduce the conditional diffusion model. In Section \ref{sec::diff_el}, we discuss the integration of Euler-Lagrange residual as a virtual observable into the diffusion model, and in section \ref{sec::bayes}, we introduce the estimation of the variance of dynamics residuals.
\subsection{Euler-Lagrange dynamics for Hand Motion}\label{sec::el}
We build on the MANO parametric model \cite{romero2017a}. Its pose parameters $\theta \in \mathbb{R}^{15\times3}$ define the rotation of 15 hand joints and its shape parameters $\beta \in \mathbb{R}^{10}$ capture the variations across hand shapes. Following established strategies~\cite{zhang2024physpt}, we assume rigid-body dynamics and a constant shape over time, and define the generalized coordinates as $\mathtt{q} = \{R, t, \theta\}$ where $R \in \mathbb{R}^{3\times3}$ and $t \in \mathbb{R}^{3}$ are the wrist rotation and translation, respectively. Using this generalized coordinates we can formulate Euler-Lagrange equation as: 
\begin{equation}
    \mathtt{M}(\mathtt{q};\mathtt{m}, \mathtt{I})\ddot{\mathtt{q}} + \mathtt{C}(\mathtt{q},\dot{\mathtt{q}};\mathtt{m},\mathtt{I}) + \mathtt{g}(\mathtt{q};\mathtt{m}) = \mathcal{F} \label{eq::el-lagrange}
\end{equation}
where $\mathtt{M}$, $\mathtt{C}$, $\mathtt{g}$ are the generalized mass matrix, Coriolis, centrifugal terms, gravitational terms, respectively ,and $\mathcal{F}$ on the right-hand side of the equation is the net generalized forces to explain observed hand motion (More explanation on these terms is provided in Appendix \ref{appx:el_terms}.). Generalized velocity $\dot{\mathtt{q}}$ and acceleration $\ddot{\mathtt{q}}$ can be obtained via the finite difference \cite{liu2012quick, xie2021physics}. To parameterize inertial properties, we obtain a mesh $\mathtt{V}_{h} \in \mathbb{R}^{778\times 3}$ via linear blend skinning function LBS \cite{romero2017a} as $\mathtt{V}_h = \text{LBS}(\theta, \beta)$ and assign each vertex to a kinematic part via MANO skinning weights. For each hand part, we compute volume (via tetrahedralization of the closed surface), mass $\mathtt{m}_k = \rho V_k, k\in \{1, \dots, 15\}$ under a constant density $\rho$ obtained from literature \cite{mirakhorlo2016anatomical} and local inertia tensor $\mathtt{I}_k$. Refer to \cite{zhang2024physpt} for further details. 

\subsection{Conditional Diffusion Model for Hand Motion Refinement}\label{sec::diffusion}
To recover physically plausible clean motion, we exploit diffusion-based framework, which supports probabilistic inference over trajectories. Specifically, we model a conditional distribution $p(\xt| \yt)$ for our final refined estimates $\xt$ conditioned on initial image-based estimates $\yt$ to approximate data distribution $q(\xt|\yt)$. 

\noindent\textbf{Forward process.} Similar to previous works~\cite{yue2023resshift, zhang2025diffusion}, we construct a Markov chain process to gradually shift from clean motion $\xt$ during training to initial motion estimates $\yt$ by adding their distance $\res = \yt - \xt$ through a shifting sequence $\{\etasche\}_{n=1}^N$ with length $N$:
\begin{equation}
    q(\xt^n|\xt^{n-1},\yt) = \mathcal{N}(\xt^n;\xt^{n-1}+\alpsche_n \res,\noi^2 \alpsche_n I)
\end{equation}
where $\noi$ is the noise variance, $\alpsche_n = \etasche_n - \etasche_{n-1}$ for $n > 1$ and $\alpsche_1 = \etasche_1$. Given this transition distribution, the marginal distribution at any diffusion step $n$ has a closed-form solution as follows:
\begin{equation}
    q(\xt^n|\xt,\yt) = \mathcal{N}(\xt^n;\xt+\etasche_n \res,\noi^2 \etasche_n I)
\end{equation}
To formulate the shifting sequence, we refer to the previous work \cite{yue2023resshift} to monotonically increase $\etasche$ from $\etasche_1 \rightarrow 0$ to $\etasche_N \rightarrow 1$. As a result, during the forward process, we start from a Dirac distribution with mean $\xt$ at diffusion step 0 and end up with a Gaussian distribution $\mathcal{N}(\yt, \noi^2 I)$ at diffusion step $N$.

\noindent\textbf{Reverse process.} During the reverse process, we compute the posterior distribution $p(\xt|\yt)$ as follows: 
\begin{flalign}
    p(\xt|\yt) = & p(\xt^N | \yt) \\ \nonumber &\int\prod_{n=1}^N   p_\phi (\xt^{n-1}|\xt^{n},\yt)d\xt^{1:N}
\end{flalign}
Here, $p(\xt^N | \yt)$ is approximately a multivariate Gaussian distribution with mean $\yt$ and variance $\noi^2$ (i.e., $\mathcal{N}(\yt, \noi^2I)$. The reverse kernel $p_\phi (\xt^{n-1}|\xt^{n},\yt)$ is the distribution we learn via a backbone with learnable weights $\phi$ which we introduce in the following subsection.

\noindent\textbf{Backbone.} Following the previous work~\cite{yue2023resshift}, we assume the reverse kernel as a Gaussian distribution, i.e., $p_\phi(\xt^{n-1}|\xt^n, \yt) = \mathcal{N}(\xt^{n-1};\mu_\phi(\xt^n,\yt,n), \Sigma_nI)$ with fixed variance as $\Sigma_n =  \frac{\etasche_{n-1}}{\etasche_{n}}\alpsche_n\noi^2$. The target reverse distribution is then formulated as follows:
\begin{equation}
    q(\xt^{n-1}|\xt^n,\xt,\yt) = \mathcal{N}(\xt^{n-1}|A_n\xt^n + B_n\xt, \Sigma_nI)\label{eq:reverse}
\end{equation}
where $A_n = \frac{\etasche_{n-1}}{\etasche_{n}}$, $B_n=\frac{\alpsche_n}{\etasche_n}$. For the derivation of Equation \ref{eq:reverse} please refer to \cite{yue2023resshift}. To estimate the mean, we reparameterize it as:
\begin{equation}
    \mu_\phi(\xt^n,\yt,n) = A_n\xt^n + B_nf_\phi(\xt^n,\yt,n)
\end{equation}
where $f_\phi(\xt^n,\yt,n)$ is the backbone with weights $\phi $ that outputs denoised motion $\hatxt$. To train the backbone, we optimize following loss function:
\begin{equation}
    \mathcal{L}_{data} = \mathbb{E}_{n\sim[1, N]}||\xt - f_\phi (\xt^n, \yt, n)||^2
\end{equation}
where $\xt$ is ground truth motion sampled from data distribution $q(\xt|\yt)$. 
\begin{figure}[t!]
    \centering
    \includegraphics[width=0.9\linewidth]{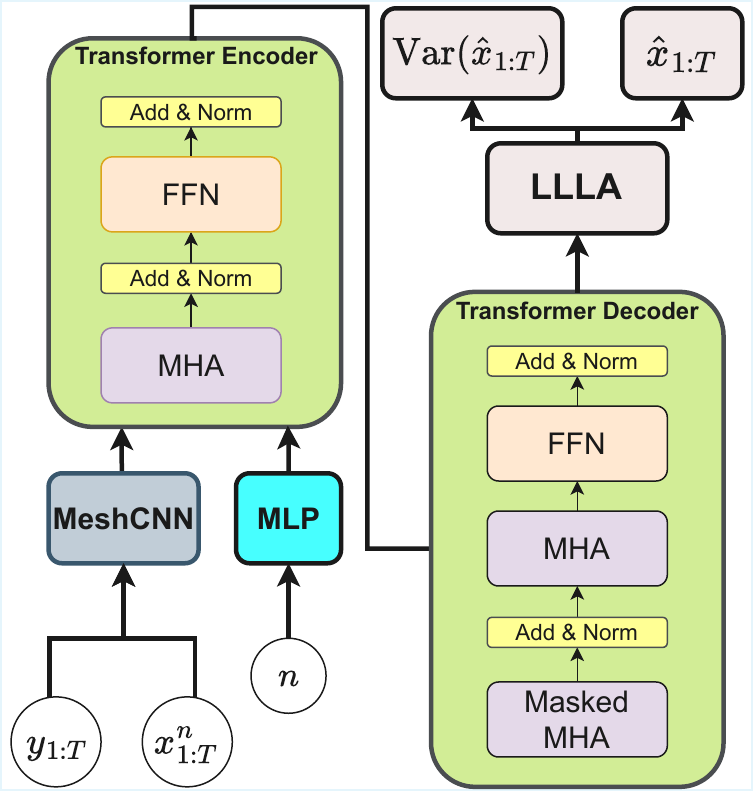}
\caption{\textbf{Backbone architecture}. At diffusion step $n$, the current pose sequence ${x}^{n}_{1:T}$ and image-based estimates ${y}_{1:T}$ are converted to meshes and encoded by MeshCNN \cite{chen2022mobrecon}, while an MLP encodes $n$. A Transformer encoder-decoder fuses these features, and an LLLA head predicts the refined pose sequence $\hat{{x}}_{1:T}$ and its variance $\mathrm{Var}(\hat{{x}}_{1:T})$.}
    \label{fig:transformer}
\end{figure}
The backbone architecture is depicted in Figure \ref{fig:transformer}. A Transformer encoder–decoder is used to model temporal dependencies across frames. We run forward kinematics with the pose and shape parameters (i.e., $\beta_{avg}$) to obtain the hand mesh sequence $\mathtt{V}_{1:T}^n$ and extract topology-aware spatial features using a MeshCNN \cite{chen2022mobrecon}. These features are concatenated with an MLP embedding of the diffusion step $n$. Up to this point, the model captures geometric and temporal patterns, but remains agnostic to motion dynamics. In the next section, we introduce a mechanism to integrate hand dynamics and render the model physics-aware.
\subsection{Physics-aware Diffusion Refinement} \label{sec::diff_el}
\noindent\textbf{Dynamic residual as a virtual observable.} 
Since the output of the backbone is a motion sequence, we can evaluate the Euler-Lagrange residual at each frame as:
\begin{equation}
    Z_t = \mathtt{M}_t\ddot{\mathtt{q}}_t + \mathtt{C}_t + \mathtt{g}_t - \hat{\mathcal{F}}_t
\end{equation}
A direct deterministic penalty would be $\mathcal{L}^{D}_{EL} = \sum_{t=1}^T||Z_t||_1$ where $\hat{\mathcal{F}}_t$ is the pseudoforce computed from motion data following~\cite{zhang2024physpt}. Rather than enforcing this loss as a hard constraint, we treat residuals as \emph{virtual observables}. This brings key benefits. The trajectories we refine come from a learned estimator that encodes strong data priors. Forcing them to exactly satisfy an approximate dynamics model (with no explicit contact force modeling, using approximate physical properties such as mass) might push model mismatch into the motion and may distort true hand dynamics. Treating residuals as virtual observations only \emph{scores} how compatible a candidate refinement is with the physics, without overriding the information already present. Therefore, we treat the residuals as virtual observables \cite{bastek2024physics, rixner2021probabilistic} sampled from the distribution below:
\begin{equation}
    q_{Z}(\zeta_{1:T} | \xt) = \mathcal{N}(\zeta_{1:T}; Z(\xt^0), \sigma^2I)
\end{equation}
where $\xt^0$  is the sample from $p(\xt|\yt)$ by performing a reverse diffusion process during training and $\sigma^2$ is a pre-defined variance.  The reason why we choose $\xt^0$ rather than $\hatxt$ is because $\hatxt$ is an estimate of $\mathbb{E}[\xt^0|\xt^n]$ and residual evaluation on this estimate is not consistent since $Z(\mathbb{E}[\xt^0|\xt^n])\neq \mathbb{E}[Z(\xt^0|\xt^n)]$ (Jensen gap \cite{gao2017bounds}). We compute the negative log likelihood loss on the residuals as:
\begin{equation}
    \mathcal{L}_{EL} =  \mathbb{E}_{n\sim[1, N]}\frac{1}{2\sigma_n}||Z_{1:T}(\xt^0)||^2
\end{equation}
where $\sigma_n = \Sigma_n/c$ is the rescaled variance for the virtual observable. 

\noindent\textbf{Total loss.} Putting everything together, our framework is trained with the following total objective:
\begin{equation}
    \mathcal{L}_{total} = \lambda_1\mathcal{L}_{data} + \lambda_2\mathcal{L}_{EL}
\end{equation}
where $\lambda_1$, $\lambda_2$ are the training weights for $\mathcal{L}_{data}$ and $\mathcal{L}_{EL}$, respectively. During training, we use the motion estimates obtained from an off-the-shelf image-based pose estimator as $\yt$ to capture the mapping distribution of the pose estimator to motion data.
\subsection{Bayesian Inference} \label{sec::bayes}
To fully capture the conditional distribution, we adopt the Last-layer Laplace Approximation (LLLA)~\cite{kou2023bayesdiff} as a post-hoc inference step. Concretely, posterior distribution on the last layer of backbone parameters is mapped to a Gaussian posterior predictive: 
\begin{flalign}
    p(\hatxt|\xt^n,n,D) \approx \mathcal{N}(&f_\phi (\xt^n, \yt, n), &\\ \nonumber &\text{diag}(\gamma^2_\phi(\xt^n, \yt, n))  
\end{flalign}
where $D = \{(\xt^s,\yt^s)\}_{s=1}^S$ is the training dataset, $\gamma^2_\phi$ is the joint-wise variance prediction function. Now we have a variance estimation for the output of the backbone but not for the final output of the diffusion model (i.e., $\xt^0)$. To achieve this, we propagate the uncertainty of the backbone by applying variance estimation on each state. Specifically, sampling for reverse diffusion is formulated as:
\begin{equation}\label{eq::sample}
    \xt^{n-1} = A_n\xt^n+B_n\hatxt+\Sigma_n\epsilon
\end{equation}
where $\epsilon \sim \mathcal{N}(0, I)$. Applying variance estimation to the both sides of equation we get:
\begin{flalign}\label{eq::var}
    \text{Var}(\xt^{n-1}) &= A_n^2\text{Var}(\xt^n)+B_n^2\text{Var}(\hatxt) \nonumber&\\&
    +\Sigma_n^2+2A_nB_n\text{Cov}(\xt^n,\hatxt)
\end{flalign}
where $\text{Var}(\hatxt) = \text{diag}(\gamma^2_\phi(\xt^n, \yt, n))$ and
\begin{flalign}\label{eq::covar}
    \text{Cov}(\xt^n,\hatxt) &\approx \frac{1}{S}\sum_{i=1}^S (\xt^{n,i}\cdot\hatxt^i)& \nonumber \\&-\mathbb{E}[\xt^n]\cdot\frac{1}{S}\sum_{i=1}^S \hatxt^i&
\end{flalign}
where $\cdot$ denotes the element-wise multiplication. Expectation of each state is then:
\begin{flalign}\label{eq::expect}
    \mathbb{E}[\xt^{n-1}] = A_n\mathbb{E}[\xt^n]+B_n\mathbb{E}[\hatxt]
\end{flalign}
For the derivations of Equations \ref{eq::var}, \ref{eq::covar} and \ref{eq::expect}, please refer to Appendix~\ref{appx::derive}. Algorithm \ref{alg:pad_hand_var} demonstrates the procedure to obtain final motion variance $\text{Var}(\xt^{0})$. We then compute the dynamics residual variance using the variance of the force inferred from motion since pseudoforce is unavailable at inference. This causes no impact because pseudoforce is only a constant shift and does not change the variance:
\begin{flalign}
    \text{Var}(\mathcal{F}_{1:T}) \approx J_{\mathcal{F}_{1:T}}\text{Var}(\xt^{0})J_{\mathcal{F}_{1:T}}^{\!\top}
\end{flalign}
where $J_{\mathcal{F}_{1:T}}=\left.\frac{\partial \mathcal{F}_{1:T}}{\partial x}\right|_{x=\mathbb{E}[\xt^{0}]}$. $\text{Var}(\mathcal{F}_{1:T})$ measures the uncertainty in the motion recovery process, enabling the identification of potentially unreliable motion estimates. 
 
\begin{algorithm}[t!]
\caption{Variance Estimation for PAD-Hand}
\label{alg:pad_hand_var}
\begin{algorithmic}[1]
\Require Starting point $x^N_{1:T} \sim \mathcal{N}(y_{1:T}, \noi^2I)$, Monte Carlo sample size $S$, Pre-trained backbone $f_\phi$.
        \Ensure  Motion generation $x^0_{1:T}$ and joint-wise uncertainty $\text{Var}(x^0_{1:T})$.
        \State Construct the joint-wise variance prediction function ${\gamma}^2_\phi$ via LLLA (Last Layer Laplace Approximation);
            \State $\mathbb{E}[x^N_{1:T}]\gets x^N_{1:T}$, $\text{Var}(x^N_{1:T})\gets \mathbf{0}$;
            \State $\text{Cov}(x^N_{1:T}, \hat x_{1:T})\gets\mathbf{0}$;
        \For {$n=N \to 1$}
            \State Sample  $\hat x_{1:T}$ from the Gaussian distribution $ \mathcal{N}({f}_{\phi}(x^n_{1:T}, n, y_{1:T}), \mathrm{diag}({\gamma}^2_\phi(x^n_{1:T}, n, y_{1:T})))$;
            \State Obtain $x^{n-1}_{1:T}$ via Equation \ref{eq::sample}; %
            \State Estimate $\mathbb{E}[x^{n-1}_{1:T}]$ via Equation \ref{eq::expect};
            \State Estimate $\text{Var}(x^{n-1}_{1:T})$ via Equation \ref{eq::var}; %
                \State Sample $x^{n-1,~i}_{1:T} \sim \mathcal{N}(\mathbb{E}[x^{n-1}_{1:T}], \text{Var}(x^{n-1}_{1:T})), i \in \{1,\dots, S$\};
            \State Estimate $\text{Cov}(x^{n-1}_{1:T},\hat x_{1:T})$ via Equation \ref{eq::covar};
        \EndFor
\end{algorithmic}
\end{algorithm}
\section{Experiments}
\subsection{Experimental Settings}
\noindent\textbf{Datasets.} To evaluate our pipeline, we employ two benchmarks which are DexYCB \cite{chao2021dexycb} and HO3D \cite{hampali2020a}. Both datasets offer high-quality per-frame 3D hand pose and shape annotations, captured with multi-view setups across diverse subjects. We adopt the official training and evaluation splits used in prior studies. 

\noindent\textbf{Metrics.}
Following prior works \cite{potamias2024wilor, pavlakos2024reconstructing, zhou2024simple, fu2023deformer, ziani2022tempclr}, we report PA-MPJPE (Procrutes aligned Mean Per Joint Error) and MPJPE (Mean Per Joint Error) on DexYCB, and PA-MPJPE (in mm) on HO3D as reconstruction accuracy metrics. For physical plausibility, we additionally report the acceleration error (ACCEL, in mm/frame$^2$).

\noindent\textbf{Implementation.} Our backbone is a Transformer encoder–decoder with four encoder and four decoder layers, each using eight attention heads and an embedding size of 512.  For per–frame spatial encoding, the MeshCNN module has four layers with channel dimensions $[32,\,64,\,64,\,64]$. Since the Transformer in  backbone is permutation-invariant, we add standard sinusoidal positional encodings to the frame-wise mesh tokens before feeding them to the Transformer.  Following prior works \cite{kocabas2020vibe,zhang2024physpt,zhang2025diffusion}, we set the sequence length to $T=16$. We train with AdamW \cite{kingma2014adam} using an initial learning rate of $2\times10^{-4}$, decayed by a factor of $0.8$ every 10 epochs. The loss weights are $\lambda_{1}=2\times10^{3}$ and $\lambda_{2}=500$. $N, S$ are set to 4 and 20, respectively. In addition, we supervise with geometric losses from \cite{tevet2022human} and include a consistency regularization term to mitigate error accumulation in diffusion sampling, following \cite{kim2023consistency} (Ablation study on these losses is provided in Appendix \ref{appx::data-driven}).

\subsection{Quantitative Results}
\begin{table}[t!] %{r}{0.6\linewidth}
    \centering
    \tabcolsep=0.01in
    \caption{\textbf{Comparison to SOTAs on DexYCB}. “*” denotes results obtained from official models; others are from original papers. ``P'' and ``D'' denote probabilistic and deterministic models, respectively.
    }
    \resizebox{\linewidth}{!}{
    \begin{tabular}{l ccccc}
    \toprule
    Method & Input& Type & PA-MPJPE$\downarrow$  & MPJPE$\downarrow$ &  ACCEL$\downarrow$  \\ \midrule
    
    MaskHand \cite{li2024hhmr} & \multirow{2}{*}{Image} & P & 5.0 & 11.70 & - \\
    *WiLoR \cite{potamias2024wilor} &  & D & 4.88 & 12.75 & 6.70 \\ \midrule
     S$^2$HAND(V)  \cite{tu2023consistent} & \multirow{5}{*}{Sequence} & \multirow{5}{*}{D} & 7.27  &   19.67 & - \\ 
    VIBE \cite{kocabas2020vibe}& &  & 6.43 & 16.95 & - \\
    TCMR \cite{choi2021beyond} & &  & 6.28 & 16.03 & - \\
    *Deformer \cite{fu2023deformer} & &  & 5.22 & 13.64 & 6.77 \\
    BioPR \cite{xie2024ms} & &  & - & 12.81 & - \\ 
    \midrule
    WiLoR + \textbf{Ours} & Sequence & P & \textbf{4.63} & \textbf{10.56} & \textbf{3.34} \\ \bottomrule
    \end{tabular}
    }
    \label{tab:main_dexycb}
\end{table}
\begin{table}[t!]
    %\begin{minipage}[t!]{0.75\linewidth}
    \centering
    \tabcolsep=0.06in
    \caption{\textbf{Comparison to SOTAs on HO3D}. “*” denotes results from official models; others are from original papers. ``P'' and ``D'' denote probabilistic and deterministic models, respectively.}
    \resizebox{\linewidth}{!}{
    \begin{tabular}{l cccc}
    \toprule
    Method & Input& Type &  PA-MPJPE$\downarrow$ &  ACCEL$\downarrow$  \\ \midrule
     
    AMVUR \cite{jiang2023probabilistic} & \multirow{2}{*}{Image}  & P  & 10.3 & - \\
    *WiLoR \cite{potamias2024wilor}  &  & D & 7.50 & 4.98 \\ \midrule
    VIBE \cite{kocabas2020vibe}& \multirow{4}{*}{Sequence} & \multirow{4}{*}{D} & 9.90 & - \\
    TCMR \cite{choi2021beyond} &  & & 11.40 & - \\
    TempCLR \cite{ziani2022tempclr} &  & &10.60 & - \\
    *Deformer \cite{fu2023deformer} &  & &9.40 & 6.37 \\
    \midrule
    WiLoR + \textbf{Ours} & Sequence & P & \textbf{7.43}  & \textbf{2.71} \\ \bottomrule
    \end{tabular}
    }
    \label{tab:main_ho3d}
    %\end{minipage}
\end{table}

\begin{figure*}[t!]
    \centering
        \includegraphics[width=0.82\linewidth]{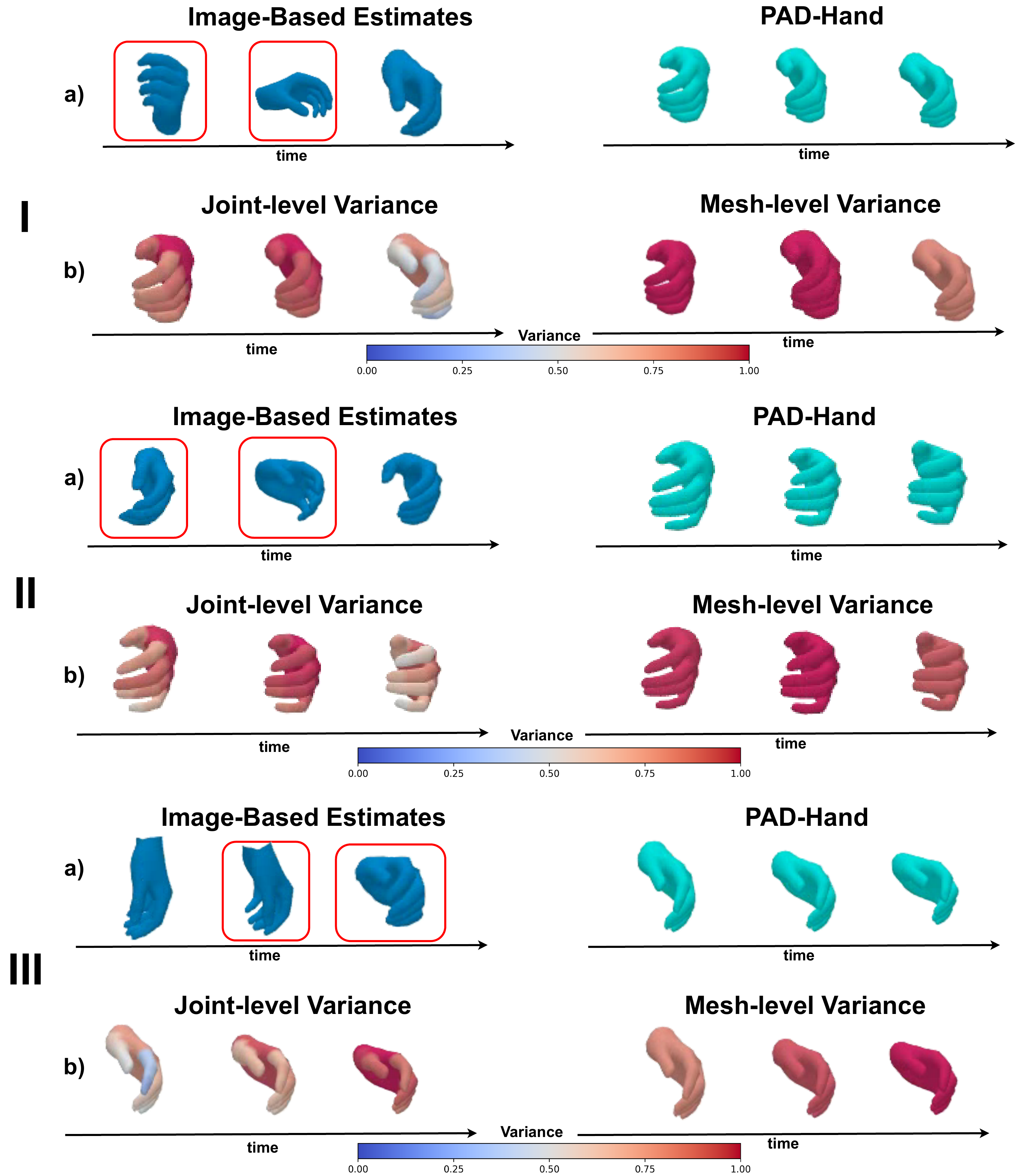}
        \caption{\textbf{Refined motion estimates by PAD-Hand with dynamic variance on DexYCB.} We visualize three representative sequences (I–III). In each block, row (a) compares the original image-based motion estimates to the trajectories refined by PAD-Hand, while row (b) shows the corresponding variance estimations in terms of joint-level and mesh-level dynamic variance. The red boxes highlight frames where the image-based estimates exhibit strong jitter.}
        \label{fig::qualitative}
\end{figure*}

\noindent\textbf{Results on DexYCB.} We initialize training with image-based estimates from the large-scale pre-trained WiLoR model \cite{potamias2024wilor} and compare our refinements against state-of-the-art video-based hand reconstruction methods. Table~\ref{tab:main_dexycb} reports results on DexYCB. Our framework improves WiLoR on both reconstruction and physical plausibility metrics: PA-MPJPE drops from $4.88$\,mm to $4.63$\,mm ($5.1\%$), MPJPE from $12.75$ mm to $10.56$\,mm ($17.2\%$), and the acceleration error from $6.70$\,mm/frame$^{2}$ to $3.34$\,mm/frame$^{2}$ ($50.1\%$). We further report improvements over two additional baselines on DexYCB in Appendix \ref{appx::more_baselines}.

\noindent\textbf{Results on HO3D.}
A similar trend is observed on HO3D. As shown in Table~\ref{tab:main_ho3d}, our framework reduces PA-MPJPE from $7.50$\,mm to $7.43$\,mm and substantially improves physical consistency, lowering the acceleration error from $4.98$\,mm/frame$^{2}$ to $2.71$\,mm/frame$^{2}$.
\subsection{Qualitative Results}
Figure~\ref{fig::qualitative} qualitatively illustrates the effect of PAD-Hand on three representative sequences from DexYCB. Across all sequences, PAD-Hand produces smoother, more physically plausible motions compared to the jittery image-based estimates. In sequence~I (a), a pronounced discontinuity in global rotation is clearly visible in the image-based motion; our refinement removes this jump and restores a coherent trajectory. The corresponding variance maps in sequence~I (b) show elevated values at both joint and mesh level around the same frames, indicating that our force variance reliably flags this problematic segment. The same pattern appears in sequences~II and~III: PAD-Hand corrects implausible pose changes while the joint-level and mesh-level variances peak exactly where the original image-based estimates are unstable. This alignment supports our claim that PAD-Hand not only improves motion quality but also provides a meaningful indicator of where the underlying estimates are unreliable. HO3D qualitative results are in Appendix \ref{appx::ho3d}.

\subsection{Ablation Study}
\noindent\textbf{Effectiveness of physics residual loss.} We validate the effectiveness of our physics loss on DexYCB. As shown in Table~\ref{tab:ablation}, integrating the Euler-Lagrange residual loss reduces error across all metrics. In terms of reconstruction accuracy, PA-MPJPE decreases from $4.65$ mm to $4.63$\,mm and MPJPE from $10.62$ mm to $10.56$\,mm. For physical plausibility, the acceleration error improves from $3.36$ mm/frame$^{2}$ to $3.34$\,mm/frame$^{2}$. Improvements are statistically significant as demonstrated in Appendix \ref{appx::statistics}, and the marginal gains are due to WiLoR's data sufficiency on this in-distribution dataset. As shown in Appendix \ref{appx::taco}, more improvements are observed when testing on an out-of-distribution dataset.
\begin{table}[ht!]
    \centering
    \caption{\textbf{Effectiveness of residual loss on DexYCB.}}
    \resizebox{\linewidth}{!}{
    \begin{tabular}{cc ccc}
    \toprule
     \multicolumn{2}{c}{Loss} & \multicolumn{3}{c}{Evaluation Metrics}   \\ \midrule
 $\mathcal{L}_{data}$  & \multicolumn{1}{c}{$\mathcal{L}_{EL}$} & \multicolumn{1}{c}{PA-MPJPE$\downarrow$} & \multicolumn{1}{c}{MPJPE$\downarrow$}&  \multicolumn{1}{c}{ACCEL$\downarrow$} \\ \midrule
  \multicolumn{2}{l}{*WiLoR \cite{potamias2024wilor}}  & 4.88 & 12.75 & 6.70 \\ \midrule
    \checkmark &   & 4.65 & 10.62 & 3.36 \\
   \checkmark & \checkmark & \textbf{4.63} & \textbf{10.56} & \textbf{3.34}
    \\
    \bottomrule
   
    \end{tabular}
    }
    \label{tab:ablation}
\end{table}

\begin{figure}[t!]
    \centering
        \includegraphics[width=\linewidth]{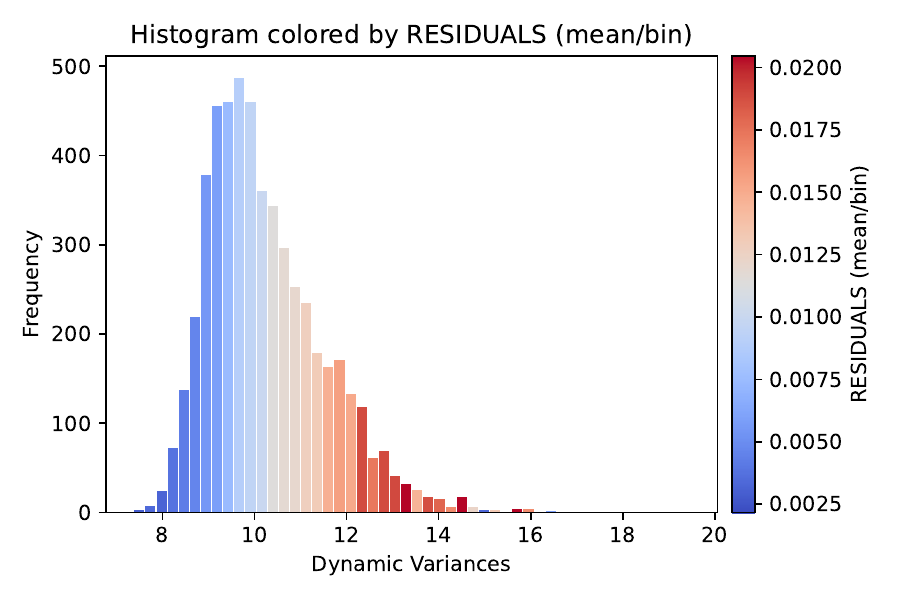}
        \caption{\textbf{Distribution of dynamic variances for PAD-Hand.} Bar color encodes the mean Euler–Lagrange residual within each variance bin (blue is low, red is high). Higher variance bins coincide with larger residuals, indicating that the model’s uncertainty aligns with physics violations.}
        \label{fig::hist}
        \vspace{-0.2cm}
\end{figure}

\noindent\textbf{Effectiveness of physics integration.} We ablate the role of physics in our probabilistic framework by comparing a deterministic penalty ($\mathcal{L}^{D}_{EL}$) with our probabilistic physics integration. As shown in Table~\ref{tab:ablation_phys}, the probabilistic variant consistently performs better, indicating that enforcing physics at the distribution level is preferable and better aligned with the probabilistic nature of diffusion models.
\begin{table}[ht!]
    \centering
    \caption{\textbf{Ablation on physics integration on DexYCB}.}
     \resizebox{0.95\linewidth}{!}{
    \begin{tabular}{l ccc} \toprule
       Loss  &  PA-MPJPE$\downarrow$ & MPJPE$\downarrow$ & ACCEL$\downarrow$\\ \midrule
        *WiLoR \cite{potamias2024wilor}  & 4.88 & 12.75 & 6.70 \\ \midrule
         $\mathcal{L}^{D}_{EL}$  & 4.66 & 10.61 & 3.35 \\ 
       \textbf{Ours} & \textbf{4.63} & \textbf{10.56} & \textbf{3.34} \\ \bottomrule
    \end{tabular}
    }
    \label{tab:ablation_phys}
    \vspace{-0.25cm}
\end{table}
\noindent\textbf{Usefulness of dynamic variance.} We visualize the estimated dynamic variances with a histogram whose bars are colored by the mean Euler–Lagrange residual within each bin. As shown in Figure~\ref{fig::hist}, higher variance bins coincide with larger residuals, indicating that increased dynamic variance flags motions that are physically implausible which is typically arising from poor image-based estimates.

\section{Conclusion}
In this paper, we introduce a physics-aware diffusion framework for 3D hand motion recovery together with its dynamic variance information. We formulate Euler--Lagrange dynamics for hand motion and treat their residuals as virtual observables in a probabilistic model, allowing physics to guide refinement without being enforced as a hard constraint. Experiments on two widely used hand datasets show consistent improvements over both image-based baselines and prior video-based methods. In addition, we propose a method to estimate per-time, per-joint dynamic variances, and demonstrate that these variances reliably highlight frames where the underlying image-based motion estimates are unreliable, providing a physically grounded indicator of motion quality. This work presents an initial use case of dynamic variances, and exploring broader applications of this information remains an interesting direction for future work.

One limitation of our approach is that the current physics modeling does not explicitly capture object geometry or contacts. We consider integrating these aspects for modeling delicate hand--object dynamics as a promising direction for future work.

\section*{Acknowledgment}
This work was supported by Michigan State University.
{
    \small
    \bibliographystyle{ieeenat_fullname}
    \bibliography{main}
}

% WARNING: do not forget to delete the supplementary pages from your submission 
\clearpage
\setcounter{page}{1}
\maketitlesupplementary

\appendix
The supplementary material is composed of an appendix and a video file.

\textbf{Appendix}
\begin{itemize}
    \item Section \ref{appx:el_terms}: Details of the Euler-Lagrange Terms
    \item Section \ref{appx::derive}: Bayesian Inference: Derivations
    \item Section \ref{appx::additional}: Additional Experiments
    \item Section \ref{appx::ho3d}: Additional Visualizations
    \item Section \ref{appx::compare_dip}: Discussion: PAD-Hand vs. DIP
\end{itemize}

\textbf{Video file}
\begin{itemize}
    \item \textbf{``Hand Motion Comparison With Variance.mp4''}: Hand motion comparison with image-based estimates from WiLoR \cite{potamias2024wilor} and includes our dynamic variance estimation visualizations at joint, mesh level. 
\end{itemize}
\section{Details of the Euler-Lagrange Terms}\label{appx:el_terms}
In this section, we further clarify the terms in Eq.~\ref{eq::el-lagrange}. Generalized mass matrix $\mathtt{M}$ captures how the hand's mass and inertia are distributed across joints. It determines how joint accelerations are related to the required generalized forces. The term $\mathtt{C}$ represents Coriolis and centrifugal effects caused by motion, accounting for velocity-dependent interactions between different joints. The term $\mathtt{g}$ corresponds to gravitational forces, i.e., the torques induced by gravity under the current hand configuration. Finally, $\mathcal{F}$ denotes the net generalized forces that explain the observed hand motion, including the forces needed to produce the measured accelerations and to balance the dynamic effects captured by $\mathtt{M}$, $\mathtt{C}$, and $\mathtt{g}$.

\section{Bayesian Inference: Derivations}\label{appx::derive}
In this section, we derive equations \ref{eq::var}, \ref{eq::covar} and \ref{eq::expect}. We start with our sampling equation
\begin{flalign}\label{appx:sampling}
    \xt^{n-1} = A_n\xt^n+B_n\hatxt+\Sigma_n\epsilon 
\end{flalign}
Taking variance on both sides leads to
\begin{flalign}
     \text{Var}(\xt^{n-1}) &= A_n^2\text{Var}(\xt^n)+B_n^2\text{Var}(\hatxt) & \nonumber \\&
    +\Sigma_n^2+2A_nB_n\text{Cov}(\xt^n,\hatxt)&\\& + 2A_nC_n\text{Cov}(\xt^n,\epsilon) + 2B_nC_n\text{Cov}(\hatxt,\epsilon) \nonumber&
\end{flalign}
Since $\epsilon \sim \mathcal{N}(0, I)$ is independent of $\xt^n$ and $\hatxt$, their covariances $\text{Cov}(\xt^n,\epsilon)$, $\text{Cov}(\hatxt,\epsilon)$ are zero. Hence we have
\begin{flalign}
     \text{Var}(\xt^{n-1}) &= A_n^2\text{Var}(\xt^n)+B_n^2\text{Var}(\hatxt) \nonumber&\\&
    +\Sigma_n^2+2A_nB_n\text{Cov}(\xt^n,\hatxt)&
\end{flalign}
Now we derive the covariance $\text{Cov}(\xt^n,\hatxt)$ as follows:
\begin{flalign}\label{appx:covar}
    \text{Cov}(\xt^n,\hatxt) = \mathbb{E}[\xt^n \cdot \hatxt] - \mathbb{E}[\xt^n]\cdot\mathbb{E}[\hatxt]
\end{flalign}
Using the law of total expectation $\mathbb{E}[\mathbb{E}[X|Y]]=\mathbb{E}[X]$, $\mathbb{E}[\hatxt]$ can be written as
\begin{flalign}\label{appx:s_term}
    \mathbb{E}[\hatxt] = \mathbb{E}_{\xt^n}[\mathbb{E}[\hatxt|\xt^n]] = \mathbb{E}_{\xt^n}[\hat X_{1:T}]
\end{flalign}
where $\hat X_{1:T} = f_\phi (\xt^n, \yt, n)$ and for the first term
\begin{flalign}\label{appx:f_term}
    \mathbb{E}[\xt^n \cdot \hatxt] &=  \mathbb{E}_{\xt^n}[\xt^n\cdot \mathbb{E}[\hatxt|\xt^n]]&\\ \nonumber&
    =\mathbb{E}_{\xt^n}[\xt^n\cdot \hat X_{1:T}]
\end{flalign}
Integrating equations \ref{appx:s_term} and \ref{appx:f_term} to the equation \ref{appx:covar} leads to
\begin{flalign}
    \text{Cov}(\xt^n,\hatxt) &= \mathbb{E}_{\xt^n}[\xt^n\cdot \hat X_{1:T}]&  \\ & - \mathbb{E}[\xt^n]\cdot\mathbb{E}_{\xt^n}[\hat X_{1:T}] \nonumber&
\end{flalign}
which we approximate via Monte Carlo (MC) estimation:
\begin{flalign}\label{appx:final_covar}
    \text{Cov}(\xt^n,\hatxt) &\approx \frac{1}{S}\sum_{i=1}^S (\xt^{n,i}\cdot\hat X_{1:T}^i)& \nonumber \\&-\mathbb{E}[\xt^n]\cdot\frac{1}{S}\sum_{i=1}^S\hat X_{1:T}^i&
\end{flalign}
where $S$ is the sample size.

For the equation \ref{eq::expect}, taking expectation on both sides of equation \ref{appx:sampling} leads to
\begin{flalign}
    \mathbb{E}[\xt^{n-1}] = A_n\mathbb{E}[\xt^n]+B_n\mathbb{E}[\hatxt] + \Sigma_n\mathbb{E}[\epsilon]
\end{flalign}
Since $\epsilon \sim \mathcal{N}(0, I)$, $\mathbb{E}[\epsilon]$ is zero and by using law of total expectation on $\mathbb{E}[\hatxt]$ we get
\begin{flalign}
    \mathbb{E}[\xt^{n-1}] = A_n\mathbb{E}[\xt^n]+B_n\mathbb{E}_{\xt^n}[\hat X_{1:T}]
\end{flalign}

\section{Additional Experiments} \label{appx::additional}
\subsection{Ablation Study on Data-driven Losses}\label{appx::data-driven} In this section, we perform an ablation study on data-driven losses which are the backbone loss $\mathcal{L}_{b}$, geometric loss $\mathcal{L}_{g}$, and consistency regularization  $\mathcal{L}_{r}$ and report the results in Table \ref{tab:data-driven}. $\mathcal{L}_{g}$ and $\mathcal{L}_{r}$ enable data-driven improvements, while further adding our physics loss yields the best performance.
\begin{table}[ht!]
    \begin{minipage}{\linewidth}
    \centering
    \caption{\textbf{Effectiveness of data-driven losses.}}
    \resizebox{\linewidth}{!}{
    \begin{tabular}{cccc ccc}
    \toprule
     \multicolumn{4}{c}{Loss} & \multicolumn{3}{c}{Evaluation Metrics}   \\ \midrule
 $\mathcal{L}_{b}$& $\mathcal{L}_{g}$& $\mathcal{L}_{r}$  & \multicolumn{1}{c}{$\mathcal{L}_{EL}$} & \multicolumn{1}{c}{PA-MPJPE$\downarrow$} & \multicolumn{1}{c}{MPJPE$\downarrow$}&  \multicolumn{1}{c}{ACCEL$\downarrow$} \\ \midrule
  \multicolumn{4}{l}{WiLoR}  & 4.88 & 12.75 & 6.70 \\ \midrule
    \checkmark & &  &   & 5.01 & 12.57 & 3.83 \\
   \checkmark & \checkmark & &  & 4.83 & 10.89 & 3.57
    \\
    \checkmark & \checkmark & \checkmark &  & 4.65 & 10.62 & 3.36 \\
    \checkmark & \checkmark & \checkmark & \checkmark  & \textbf{4.63} & \textbf{10.56} & \textbf{3.34} \\
    \bottomrule
    \end{tabular}}
    \label{tab:data-driven}
    \end{minipage}
\end{table}
\subsection{Refining Additional Baselines and Robustness to Initial Pose Estimates}\label{appx::more_baselines}
Since \ours~takes the predictions of a baseline model as input and refines them, it can be applied on top of different hand pose estimators. It is also robust to poor initial poses. To demonstrate this generality, we further evaluate \ours~using the outputs of two recent baseline models, as shown in Tab.~\ref{tab::more_baselines}. In both cases, \ours~consistently improves PA-MPJPE, MPJPE, and ACCEL, showing that our method serves as an effective refinement module across different baselines. We further test robustness by replacing 80\% of a sequence with Gaussian noise, which increases the initial PA-MPJPE to 24.27. Under this severe corruption, our model reduces it to 6.53.
\begin{table}[ht!]
    \begin{minipage}{\linewidth}
    \centering
    \caption{\textbf{Comparison to more SOTAs on DexYCB.}}
    \resizebox{\linewidth}{!}{
    \begin{tabular}{l cccc}
    \toprule
    Method &  PA-MPJPE$\downarrow$ & MPJPE$\downarrow$  & ACCEL$\downarrow$  \\ \midrule
    HaMeR \cite{pavlakos2024reconstructing}  & 4.70 & 16.71 & 7.14 \\
    HaMeR+\textbf{Ours}  & \textbf{4.60}   & \textbf{11.09} & \textbf{3.37} \\ \midrule
    HandOccNet \cite{park2022a} & 5.80 & 14.0 & 9.03 \\
    HandOccNet+\textbf{Ours}  & \textbf{5.31} & \textbf{10.95}  & \textbf{3.38} \\ \bottomrule
    \end{tabular}
    }
    \label{tab::more_baselines}
    \end{minipage}
\end{table}
\subsection{Statistical Significance Analysis} \label{appx::statistics}
In this section, we verify that the performance gains obtained with $\mathcal{L}_{EL}$ are statistically significant.  To this end, we train two model variants on DexYCB using 5 random seeds: one with $\mathcal{L}_{EL}$ and one without it. We then report the mean and standard deviation over the evaluation metrics. \ours~achieves $4.61 \pm 0.02$, $10.57 \pm 0.02$, and $3.33 \pm 0.01$ for PA-MPJPE, MPJPE, and ACCEL, respectively, compared with $4.66 \pm 0.01$, $10.65 \pm 0.02$, and $3.36 \pm 0.01$ for the variant without the physics residual. These improvements are statistically significant under a paired $t$-test, with $p$-values of $0.001$, $0.001$, and $0.002$, respectively.

\subsection{Ablation Study on a Challenging Dataset}\label{appx::taco}
In this section, we evaluate the generalization ability of PAD-Hand and significance of $\mathcal{L}_{EL}$ by testing it on a dataset that was not used during training. For this, we employ a more challenging hands–object interaction dataset TACO~\cite{liu2024taco}, which contains diverse and complex manipulation sequences. The quantitative results in Table~\ref{tab::taco} show that PAD-Hand reduces PA-MPJPE from 8.37 mm to 8.02 mm and MPJPE from 25.13 mm to 24.38 mm. In addition, we observe a substantial improvement in acceleration error, which decreases from 5.47 mm/frame$^2$ to 1.84 mm/frame$^2$, indicating that our method produces smoother and more physically plausible motions in this challenging setting. Furthermore, using $\mathcal{L}_{EL}$ shows stronger generalization, improving PA-MPJPE by 0.35 compared to 0.13 with $\mathcal{L}_{data}$ alone (Tab.~\ref{tab::taco}), ensuring physically grounded motion beyond the training distribution. In addition, when trained with only 10\% of the DexYCB training data, adding this loss during WiLoR refinement effectively reduces performance degradation as shown in Table \ref{tab:less_data}. The PA-MPJPE is 6.67 with $\mathcal{L}_{EL}$ and degrades to 7.90 without $\mathcal{L}_{EL}$, demonstrating $\mathcal{L}_{EL}$'s value as a physical regularizer under limited data. 

\begin{table}[ht!]
    \begin{minipage}[t!]{\linewidth}
    \centering
\caption{\textbf{Generalization to unseen TACO~\cite{liu2024taco}.} S1 testing split is used for evaluation.} 
    \resizebox{\linewidth}{!}{
    \begin{tabular}{lccc}
    \toprule
    Method  & PA-MPJPE$\downarrow$  & MPJPE$\downarrow$ &  ACCEL$\downarrow$  \\ \midrule
    WiLoR  & 8.37 & 25.13 & 5.47 \\  \midrule
    $\mathcal{L}_{data}$  & 8.24 & 25.02 & 2.44 \\ 
     $\mathcal{L}_{data}$ + $\mathcal{L}_{EL}$ (\textbf{Ours}) &\textbf{8.02} & \textbf{24.38} & \textbf{1.87} \\ \bottomrule
    \end{tabular} 
    }
    \label{tab::taco}
    \end{minipage}
\end{table}
\begin{table}[ht!]
    \begin{minipage}[t!]{\linewidth}
    \centering
    \tabcolsep=0.2in
    \caption{\textbf{Data efficiency under $\mathcal{L}_{EL}$ on DexYCB.}}
    \resizebox{\linewidth}{!}{
    \begin{tabular}{l ccc}
    \toprule
    Method &  PA-MPJPE$\downarrow$ & MPJPE$\downarrow$   \\ \midrule
    $\mathcal{L}_{data}$  & 7.90  & 27.52 \\
    $\mathcal{L}_{data}$ + $\mathcal{L}_{EL}$ (\textbf{Ours})  & \textbf{6.67} & \textbf{25.84} \\ \bottomrule
    \end{tabular}
    }
    \label{tab:less_data}
    \end{minipage}
\end{table}

\subsection{Additional Results on Physical Plausibility} \label{appx::el}
In this section, we further assess the effectiveness of our physics integration by computing the Euler-Lagrange residual as an additional metric of physical plausibility,  defined as
\begin{flalign}
    \mathcal{R}(\mathtt{q}) = \frac{1}{T}\sum _{t=1}^T  ||\mathtt{M}_t\ddot{\mathtt{q}} + \mathtt{C}_t + \mathtt{g}_t - \mathcal{\bar F}||_1
\end{flalign}
where $\mathcal{\bar F}$ is the pseudoforce computed from ground-truth motion data. Unlike the acceleration error, which primarily captures temporal smoothness, $\mathcal{R}(\mathtt{q})$ directly measures consistency with our physics formulation and therefore offers a complementary perspective on motion quality. As reported in Table~\ref{appx:ablation_el}, our physics-aware integration achieves a lower residual than the deterministic counterpart, indicating that it produces motions that are more consistent with the underlying dynamics. Also, deterministic integration results in worse performance in PA-MPJPE (4.66 mm vs. 4.65 mm) in comparison to the data-driven approach (i.e., $\mathcal{L}_{data}$). For reference, we also include results for the SmoothFilter \cite{young1995recursive} which applies Gaussian smoothing to the WiLoR's predictions and can be viewed as a naive baseline that implicitly drives high-frequency temporal variations (e.g., acceleration) toward zero, without modeling uncertainty in the underlying image-based estimates. Notably, although SmoothFilter outperforms PAD-Hand in terms of acceleration error, reconstruction accuracy is far below than us indicating the need to have $\mathcal{R}(\mathtt{q})$ metric for a better comparison in terms of physical plausibility.
\begin{table}[t!]
    \centering
    \tabcolsep=0.03in
    \caption{\textbf{Ablation on physics integration on DexYCB}. $\dagger$ denotes that values are scaled by $10^{-3}$.}
     \resizebox{\linewidth}{!}{
    \begin{tabular}{l cccc} \toprule
       Loss  &  PA-MPJPE$\downarrow$ & MPJPE$\downarrow$ & ACCEL$\downarrow$ & $ ^\dagger \mathcal{R}(\mathtt{q})\downarrow$\\ \midrule
        *WiLoR \cite{potamias2024wilor}  & 4.88 & 12.75 & 6.70 & 19.12\\ \midrule
        SmoothFilter \cite{young1995recursive}  & 4.77 & 12.82 & \textbf{3.10} & 14.33\\ \midrule
          $\mathcal{L}_{data}$  & 4.65 & 10.62 & 3.36 & 15.16 \\ 
        $\mathcal{L}_{data}$ + $\mathcal{L}^{D}_{EL}$  & 4.66 & 10.61 & 3.35 & 13.45\\ 
       \textbf{Ours} & \textbf{4.63} & \textbf{10.56} & 3.34 & \textbf{9.04} \\ \bottomrule
    \end{tabular}
    }
    \label{appx:ablation_el}
\end{table}

\begin{figure*}[ht!]
    \centering
    \includegraphics[width=0.92\linewidth]{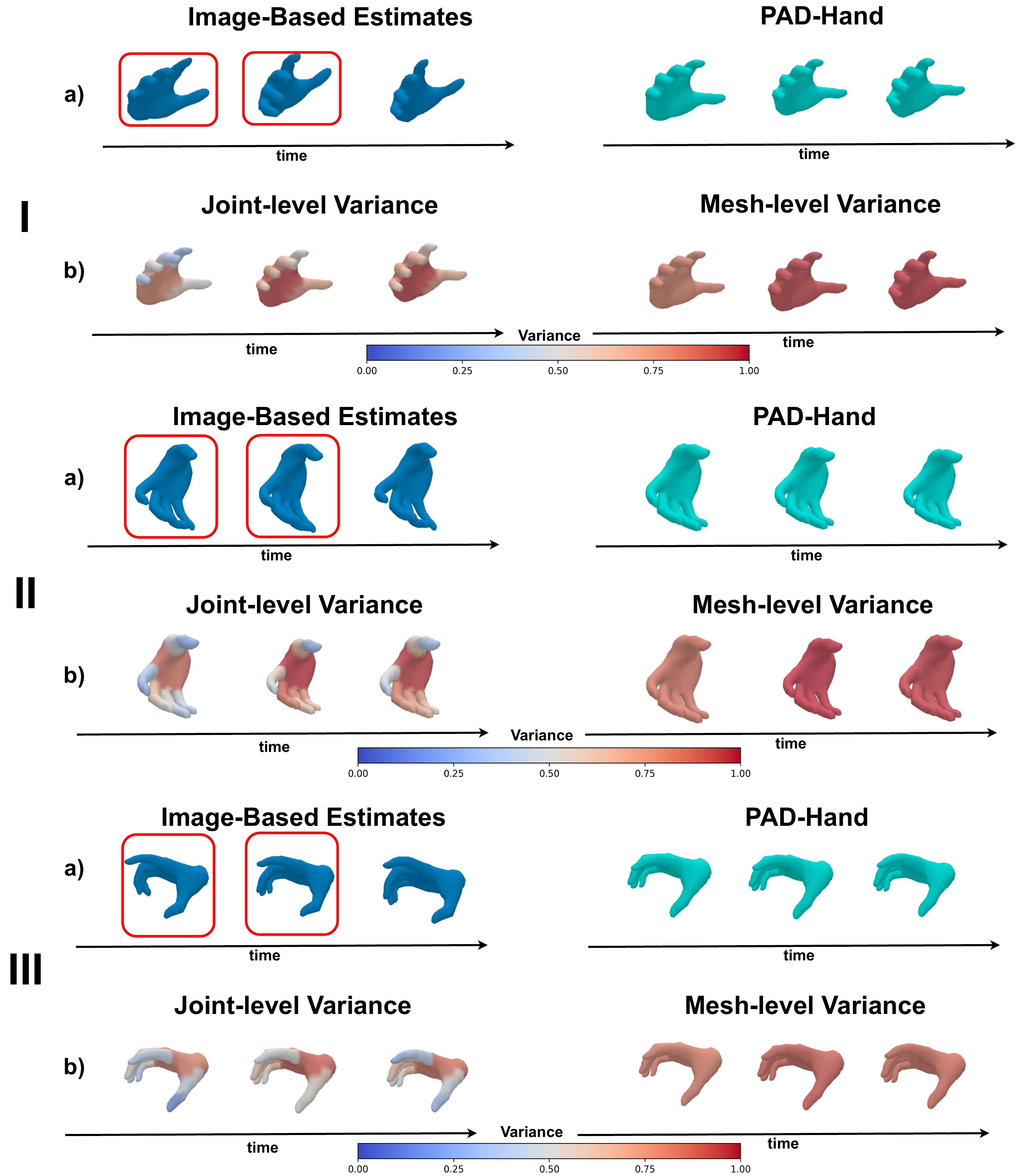}
        \caption{\textbf{Refined motion estimates by PAD-Hand with dynamic variance on HO3D.} We visualize three representative sequences (I–III). In each block, row (a) compares the original image-based motion estimates to the trajectories refined by PAD-Hand, while row (b) shows the corresponding variance estimations in terms of joint-level and mesh-level dynamic variance. The red boxes highlight frames where the image-based estimates exhibit strong jitter.}
        \label{fig::qualitative_supp}
\end{figure*}
\section{Additional Visualizations}\label{appx::ho3d}
In this section, we present additional visualizations on three representative HO3D sequences to further illustrate the effectiveness of PAD-Hand (Figure \ref{fig::qualitative_supp}). Across all examples, our pipeline not only refines the jittery motions produced by the baseline but also reliably flags segments with rapid changes through elevated variance estimates. For instance, in sequence~III (a), there is a sudden change in the thumb motion and the corresponding variance maps in sequence~III (b) assign a higher variance to the thumb joint, indicating an increased dynamic uncertainty in this fast-motion regime. Similar patterns are observed in the other sequences, where motion discontinuities and high-frequency jitter are both smoothed in the refined trajectories and highlighted by increased variance, providing an interpretable indicator of challenging image-based estimates. We also provide video versions of these sequences in ``Hand Motion Comparison With Variance.mp4'' file.

\section{Discussion: PAD-Hand vs. DIP}\label{appx::compare_dip}
Closest to our work is the DIP model proposed by Zhang \textit{et al.}~\cite{zhang2025diffusion}, which also employs a conditional diffusion model for hand motion recovery. However, our approach differs from DIP in several important aspects:
\begin{itemize}
    \item DIP is trained with synthetic noise by adding Gaussian noise to the ground-truth motion, whereas PAD-Hand takes the predictions of an image-based estimator as input during training.
    \item DIP additionally outputs hand shape parameters and is supervised with a shape loss, while in our pipeline the shape parameters are kept fixed and we focus on refining the pose parameters.
\end{itemize}
Because of these design differences, a strictly fair comparison is not possible. Nevertheless, we report DIP results and use their officially released models for evaluation. Note that evaluating $\mathcal{R}(\mathtt{q})$ on HO3D is not feasible, since pose parameters for the test set are not publicly available. As shown in Tables~\ref{appx:ablation_dip_dexycb} and~\ref{appx:ablation_dip_ho3d}, PAD-Hand consistently outperforms DIP in terms of reconstruction accuracy. Although DIP achieves a lower acceleration error on DexYCB (Table~\ref{appx:ablation_dip_dexycb}), its Euler-Lagrange residual $\mathcal{R}(\mathtt{q})$ remains higher than ours, indicating that our trajectories are better aligned with the underlying dynamics.
\begin{table}[ht!]
    \centering
    \tabcolsep=0.03in
    \caption{\textbf{Comparison to DIP \cite{zhang2025diffusion} on DexYCB}. $\dagger$ denotes that values are scaled by $10^{-3}$.}
     \resizebox{\linewidth}{!}{
    \begin{tabular}{l cccc} \toprule
       Loss  &  PA-MPJPE$\downarrow$ & MPJPE$\downarrow$ & ACCEL$\downarrow$ & $ ^\dagger \mathcal{R}(\mathtt{q})\downarrow$\\ \midrule
        *WiLoR \cite{potamias2024wilor}  & 4.88 & 12.75 & 6.70 & 19.12\\ \midrule
        WiLoR + DIP \cite{zhang2025diffusion}  & 4.81 & 12.88 & \textbf{3.23} & 13.22\\ \midrule
       WiLoR + \textbf{Ours} & \textbf{4.63} & \textbf{10.56} & 3.34 & \textbf{9.04} \\ \bottomrule
    \end{tabular}
    }
    \label{appx:ablation_dip_dexycb}
\end{table}
\begin{table}[ht!]
    \centering
    \tabcolsep=0.2in
    \caption{\textbf{Comparison to DIP \cite{zhang2025diffusion} on HO3D}.}
     \resizebox{\linewidth}{!}{
    \begin{tabular}{lcc} \toprule
       Loss  &  PA-MPJPE$\downarrow$ & ACCEL$\downarrow$\\ \midrule
        *WiLoR \cite{potamias2024wilor}  & 7.50 & 4.98 \\ \midrule
        WiLoR + DIP \cite{zhang2025diffusion}  & 7.55  & 3.03 \\ \midrule
       WiLoR + \textbf{Ours} & \textbf{7.43} & \textbf{2.71} \\ \bottomrule
    \end{tabular}
    }
    \label{appx:ablation_dip_ho3d}
\end{table}

\end{document}